\documentclass[conference]{IEEEtran}
\IEEEoverridecommandlockouts
\usepackage{cite}
\renewcommand{\baselinestretch}{0.95} 
\usepackage{amsmath,amssymb,amsfonts}
\usepackage{graphicx}
\usepackage{textcomp}
\usepackage{xcolor}
\usepackage{enumitem}
\usepackage{booktabs}
\usepackage{multirow}
\usepackage{tikz}
\usepackage{pgfplots}
\usepackage{pgfplotstable}
\usepackage{tikz}
\usepackage{pgfplotstable}
\usepackage{amsmath,amssymb,amsfonts}
\usepackage{graphicx}
\usepackage{textcomp}
\usepackage{xcolor}
\usepackage{svg}
\usepackage{subcaption}

\usepackage[ruled,vlined,linesnumbered,noresetcount]{algorithm2e}
\SetKwFor{For}{for (}{) $\lbrace$}{$\rbrace$}

\usepackage{algpseudocode} 
\usepackage{siunitx,booktabs} 
\usepackage{color,soul} 
\usepackage{xcolor}
\usepackage{flushend}
\usepackage{graphicx}
\usepackage{amssymb}
\usepackage{amsmath}
\usepackage{pdfpages}
\usepackage{lineno}
\usepackage{booktabs}
\usepackage{pifont}
\usepackage{tabu}
\usepackage{rotating}
\usepackage{array}
\usepackage[acronym,nomain]{glossaries}       
\usepackage{xcolor}
\usetikzlibrary{intersections,calc}
\usepackage{pgfplots}
\usepackage{breqn}
\usepackage{booktabs}
\usepackage{soul}
\usepackage{tikz}
\usepackage{amsmath,amssymb,amsfonts}
\usepackage{graphicx}
\usepackage{colortbl}
\usepackage{multirow}
\usepackage[numbers,super]{}
\usepackage{pgf}
\usepackage{tikz,pgfplots}

\usepackage[ruled,vlined,noresetcount]{algorithm2e}
\usepackage[linesnumbered,ruled,vlined]{algorithm2e}
\usepackage{pgfplots}
\usepackage{pgfplotstable}
\usepackage{tikz}
\usepackage{amsmath,amssymb,amsfonts}
\usepackage{graphicx}
\usepackage{textcomp}
\usepackage{xcolor}
\usepackage{svg}
\usepackage{hyperref}
\usepackage[textwidth=3.1em]{todonotes}
\usepackage{tikz}
\usepackage{subfiles}
\usepackage{subcaption}
\usetikzlibrary{patterns}
\begin{document}

\title{AMSER: \underline{A}daptive \underline{M}ultimodal \underline{S}ensing for Energy \underline{E}fficient and \underline{R}esilient eHealth Systems}

\author{\IEEEauthorblockN{Emad Kasaeyan Naeini\textsuperscript{1}, Sina Shahhosseini\textsuperscript{1}, Anil Kanduri\textsuperscript{2}, Pasi Liljeberg\textsuperscript{2}, Amir M. Rahmani\textsuperscript{1}, Nikil Dutt\textsuperscript{1}}
\IEEEauthorblockA{\textit{\textsuperscript{1} Dept. of CS, University of California, Irvine, USA, \textsuperscript{2} Dept. of Computing, University of Turku, Finland} \\
ekasaeya@uci.edu, sshahos@uci.edu, spakan@utu.fi, pakrili@utu.fi, a.rahmani@uci.edu, dutt@uci.edu}
}

\maketitle

\begin{abstract}
eHealth systems deliver critical digital healthcare and wellness services for users by continuously monitoring physiological and contextual data. eHealth applications use multi-modal machine learning kernels to analyze data from different sensor modalities and automate decision-making. Noisy inputs and motion artifacts during sensory data acquisition affect the i) prediction accuracy and resilience of eHealth services and ii) energy efficiency in processing garbage data. Monitoring raw sensory inputs to identify and drop data and features from noisy modalities can improve prediction accuracy and energy efficiency. We propose a closed-loop monitoring and control framework for multi-modal eHealth applications, AMSER, that can mitigate garbage-in garbage-out by 
i) monitoring input modalities,
ii) analyzing raw input to selectively drop noisy data and features, and
iii) choosing appropriate machine learning models that fit the configured data and feature vector - to improve prediction accuracy and energy efficiency. We evaluate our AMSER approach using multi-modal eHealth applications of pain assessment and stress monitoring over different levels and types of noisy components incurred via different sensor modalities. 
Our approach achieves up to 22\% improvement in prediction accuracy and 5.6$\times$  energy consumption reduction in the sensing phase against the state-of-the-art multi-modal monitoring application. 

\end{abstract}


\section{Introduction}
\label{intro}


eHealth applications deliver digital healthcare and wellness services for users by continuously monitoring and analyzing physiological and contextal sensory data \cite{o2011role}. 
eHealth applications feature: 
i) high volumes of input data from multi-modal sensory devices, 
ii) intense machine learning algorithmic computations for data analysis, and 
ii) limited energy and compute resources, particularly at the sensing layer (e.g., wearable sensors) \cite{naeini2019edge}. 
A large category of eHealth applications are designed to monitor users in \textit{everyday settings} through wearable sensors, acquiring often noisy continuous longitudinal data, compared to that of reliable clinical-grade sensors used in clinical settings \cite{yang2014biosensor}. 
Input sensory data perturbations including noisy and unreliable signal components, motion artifacts, and physical failure of sensors challenges the resilience of eHealth applications \cite{naeini2019PPGQA}. 
Processing perturbed sensory inputs affects prediction accuracy of machine learning algorithms, while incurring significant computational and energy expense \cite{multimodal_ehealth}. 

Multi-modal eHealth applications are executed in two phases viz., \textit{sensing} - to acquire complementary data from multi-modal sensors for an application, and \textit{sense-making} - to analyze the acquired data for predictive results \cite{mmdl-root}. Input data perturbations at the multi-modal sensing phase affects the computational workload and energy consumption in the subsequent sense-making phase. Selective sensing through data filtering and pre-processing reduces the total data volume to be sensed and processed at the sensing phase itself which consequently leads to energy savings. Selective sensing extracts insightful data to alleviate compute pressure, reduce energy consumption and improve prediction accuracy in the sense-making phase. State-of-the-art multi-modal applications use selective sensing techniques such as data filtering and compression \cite{selective_sensing->compression}, context-aware sensing \cite{selective_sensing->context_aware_sensing}, and heterogeneity-aware sensing \cite{selective_sensing->hetero_sensing} to reduce the total input data volume. Other techniques target model optimization to reduce the compute intensity of sense-making algorithms \cite{amiri2020context}. Existing optimizations for multi-modal machine learning (MMML) based eHealth applications address multi-modal sensing and sense-making independently \cite{9018161, milets, liu2018demand}. Improving end-to-end system metrics of performance, energy consumption, and prediction accuracy necessitates \textit{joint optimization of sensing and sense-making phases}. This exposes opportunities for i) adaptive sensing: selectively extracting quality-related input data and features - to alleviate compute pressure and reduce energy consumption incurred in processing garbage and/or less insightful data, and ii) adaptive sense-making: choosing suitable machine learning models by considering selective features and modalities from the sensing phase - to improve prediction accuracy. Joint optimization of multi-modal sensing and sense-making requires continuous monitoring of sensor modalities to identify input data perturbations, selective feature aggregation to isolate quality-related inputs, and choice of learning algorithms suitable for the given sensor modalities and feature vectors. 
To this end, we propose an adaptive multi-modal sensing and sense-making approach, AMSER, for improving performance, energy consumption, and prediction accuracy of eHealth applications. 
To demonstrate the potential of our AMSER approach in improving both sensing and sense-making, we design a proof-of-concept rule-based controller for adaptive feature selection, machine learning model selection, and sensor configuration, based on the monitored signal qualities. 
We integrate the controller into our framework running real eHealth applications to guide the sensing and sense-making joint optimization decisions in real-time for resilient, low-latency, and energy efficient eHealth services.
We summarize our contributions as follows:
\begin{itemize}[leftmargin=*]
    \item Design of sensor-edge framework for multi-modal machine learning based eHealth applications, capable of monitoring input signal quality, and detection of discrepancies to guide sensing and sense-making optimization decisions.  
    \item Design of a proof-of-concept rule-based controller for adaptive feature selection, modality selection, machine learning model selection, and sensor configuration based on sensor modality monitoring - to improve performance, energy, and prediction accuracy.
    \item Demonstration of our framework's efficiency on two eHealth case studies:  pain assessment and stress monitoring. 
\end{itemize}
\section{Motivation}

\begin{figure}[tb!]
    \centering
     \includegraphics[width=0.49\textwidth]{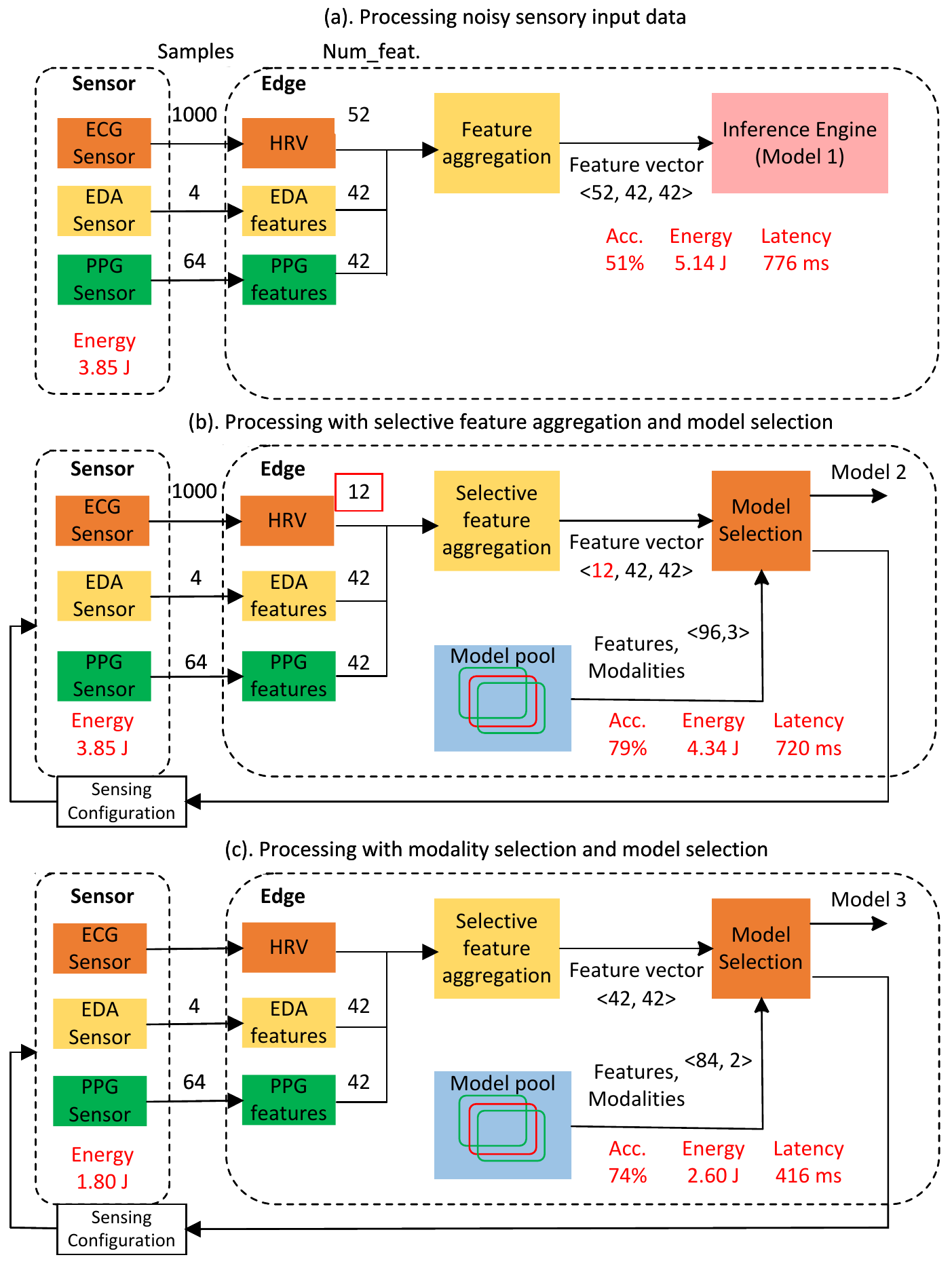}
      \vspace{-6mm}
      \caption{Example experimental scenario demonstrating adaptive sensing configuration for feature and modality selection.}
          \vspace{-6mm}
      \label{fig.motivation}
\vspace{-3mm}
\end{figure}

We illustrate the resilience challenges in MMML-based eHealth services and the significance of adaptive sensing and control in addressing those challenges through a motivational example of a pain monitoring application \cite{werner2019automatic}. The pain monitoring application acquires physiological data from different modalities viz., Electrocardiography (ECG), Electrodermal Activity (EDA), and Photoplethysmography (PPG) sensors to capture the autonomic nervous system activity against pain. Figure \ref{fig.motivation} shows the pipeline composed of multi-modal sensor data acquisition, feature aggregation, and machine learning model selection - for training and prediction. In this example, sensors from each of the ECG, EDA, and PPG modalities are sampled at 500Hz (2 channels), 4Hz, and 64 Hz, respectively. We extract relevant features from the raw input data and aggregate the features from all the modalities into an early-fused feature vector. We select a suitable machine learning model and train using the aggregated features to predict the pain levels. The model yields an average accuracy of 81\% considering no noise on the input modalities. In practical scenarios, sensory data from one or more modalities can be noisy, and/or have data unavailability with perturbations such as motion artifacts, physical damages and battery shutdown \cite{naeini2019edge}. In this example, we describe the implications of such sensory data perturbations on prediction accuracy and energy consumption, using different approaches of application orchestration.\\
\noindent\textbf{(a) Processing noisy data.} Figure \ref{fig.motivation} (a) shows the baseline scenario of processing sensory data without considering the noisy components. The ECG, EDA, and PPG modalities generate data of 8KB, 32B, and 512B, with 52, 42, and 42 features, respectively. 
In this case, the data from the ECG modality has a significant noisy component (e.g., due to improper patch contact to the chest), while the EDA and PPG modalities (e.g., collected from the wrist) generate quality inputs. 
Ignoring the input quality, features from all the modalities are fused into a vector comprising 136 features. 
The model pool comprises a set of machine learning models suitable for different feature vectors and input modalities. 
From the model pool, a machine learning model that suits the feature vector fused from 3 modalities (3-modal, 136 feature) is chosen for training and prediction. 
This model yields an average accuracy of 51\%, with an end-to-end latency of 776 ms, and energy consumption of 5.14 J and 3.85 J, for the edge and the sensor devices, respectively. 
The loss in prediction accuracy can be attributed to the noisy data from the ECG modality (garbage-in garbage-out effect). 
Further, this levies an unnecessary energy consumption incurred in sensing and processing the noisy data.
\begin{figure*}[tb!]
\centering
    \includegraphics[width=\textwidth]{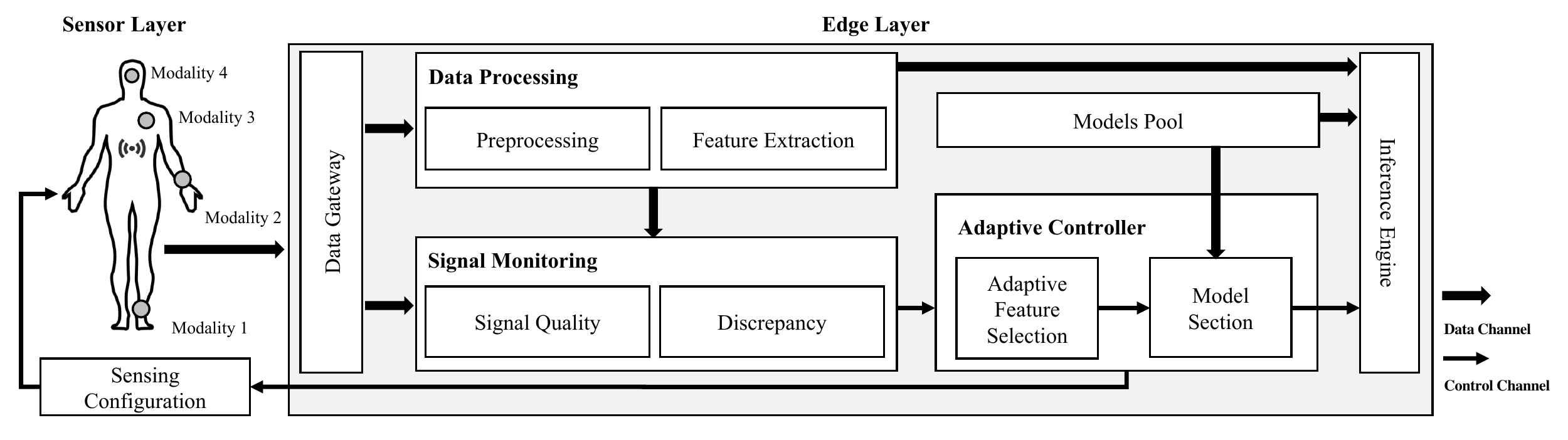}
    \vspace{-8mm}
    \caption{System Architecture Overview.}
    \label{fig:overview}
\vspace{-6mm}
\end{figure*}

\noindent\textbf{(b) Processing with selective feature aggregation.} Figure \ref{fig.motivation} (b) shows an optimized scenario with selective feature aggregation for handling modalities with noisy data. 
By considering the noisy components and aggregating only a selected set of features from ECG modality, we improve prediction accuracy, while lowering the computational effort and energy consumption. 
In this scenario, the \textit{Selective feature aggregation} selects 12 features from ECG modality (shown in red box) among the original 52 full scale features, and aggregates them with EDA and PPG features. The updated feature vector is sent to the \textit{Model Selection} to choose a machine learning model that suits the feature vector for training and prediction. This model with selective feature aggregation yields an average accuracy of 79\%, which is 35\% higher than that of the baseline scenario with no input data considerations and only 2\% less than the baseline with the no-noisy modality condition (ideal condition). Reduced features from the noisy ECG modality lowers the computational effort and energy consumption. In this case, selective feature aggregation improved the latency by 7\% and energy consumption of the edge device by 16\%. The sensing energy consumption does not change in this scenario since sensing configuration is unchanged. \\
\noindent\textbf{(c) Processing with modality selection:} Figure \ref{fig.motivation} (c) shows the optimized scenario with modality selection for handling modalities with unavailable data. In this scenario, data from the ECG modality is completely unavailable. With data from the EDA and PPG modalities being available, features from these modalities can be aggregated to build an updated feature vector. The \textit{Selective feature aggregation} aggregates 42 features from EDA and 42 features from PPG modalities. From the \textit{Models Pool}, the \textit{Model Selection} chooses a machine learning model that suits the updated feature vector with reduced number of modalities and features (2-modal, 84 feature) for training and prediction. This model yields an average accuracy of 74\%, which is 31\% higher than that of the baseline scenario with no input data considerations. 
Note that reducing the number of modalities to 2 (from 3 in Scenario (b)) lowers the accuracy only by 7\% (compared to the ideal scenario). Dropping the unavailable ECG modality reduces the total data volume to be processed significantly. If the information on the sense-making decision is fed back to the sensing layer to adaptively change the sensing configuration, further energy saving at the sensing layer is also possible, which is critical in battery-powered wearable sensors. 
Overall, this leads to 46\% improved latency, 50\% lower energy consumption at the edge device, and 53\% lower energy consumption at the sensors as compared to the baseline scenario of processing noisy data.  

The example scenarios presented in Figure \ref{fig.motivation} demonstrates the advantages of monitoring input modalities for selective feature aggregation, modality selection, and model selection. However, this requires continuous monitoring of input modalities, and intelligent control for selective aggregation, modality, and model selection. 
To this end, we propose an automated framework for adaptive multi-modal sensing to improve resilience and energy efficiency of MMML-based eHealth applications.

\section{Adaptive Multimodal Sensing-SenseMaking Framework}
\label{AMS_framework}
We deploy a commonly used eHealth system architecture to implement our AMSER approach, that consists of multiple sensor devices at the sensor layer and computing resources at the edge layer. 
Figure \ref{fig:overview} shows an overview of the proposed framework. The \textit{sensor layer} includes the multi-modal signal capability needed for eHealth monitoring applications. The edge layer is a computing device providing data gateway, data processing, signal monitoring, adaptive control, and inference at the edge. 
We detail each level of the framework below:



\noindent\textbf{Data Gateway} receives raw signal data from multi-modal sensors. The physiological signals can potentially contain two main types of noises: Baseline Wander (BW) and Motion Artifacts (MA). 

\noindent\textbf{Data Processing} performs pre-processing and feature extraction with the data collected from multi-modal sensors. 
\textit{Pre-processing} module consists of data synchronization of multi-modal signals, and bio-signal processing (e.g., peak detection) from different modalities such as ECG, PPG, and EDA 
that are essential for feature extraction, and finally filtering sensor modality inputs to produce clean signals. \textit{Feature Extraction} module extracts informative and non-redundant values from the filtered signals. We extract time domain (mean, standard deviation, rms) and frequency domain (power spectral density, median frequency, central frequency) handcrafted features among the automatic features using a variational autoencoder for all modalities
\cite{pyEDA}. 
This process facilitates subsequent learning, leading to better interpretations for signal quality assessments. 

\noindent\textbf{Signal Monitoring} handles quality assurance of the data and features to be used in the inference engine. This module observes the system parameters, disruptive events, data quality, and control flow to analyze the situation and context. 
\textit{Signal Monitoring} assesses data and its extracted features quality by monitoring key parameters from the sensing phase to identify events and triggers for joint optimization of sensing and sense-making. 
For example, a disruptive event in input data of a specific modality (e.g., motion artifacts or sensor detachment) is a trigger at the sensing phase. 
In this regard, as shown in Algorithm \ref{alg:sm}, signal monitoring assesses data ($D_M^i$) and its extracted feature ($F_M^i$) quality and outputs the level of reliability of each modality ($Rel_M^i$). 

\noindent\textit{Signal quality}: We consider three levels of signal quality viz., reliable, noisy, and uncertain. \textit{Signal quality} will assess the quality of sensing in terms of electrodes attachment, signal-to-noise ratio (SNR), and motion artifacts to determine the level of reliability.  A signal with no noisy components is labeled as \textit{reliable}. Signals extracted from sensor modalities whose electrodes are detached from the subject's body and signals with an SNR lower than a fixed threshold \cite{signal_quality->ecgppg} are labeled as \textit{noisy}. Signals with an SNR above the fixed threshold yet below the acceptable levels are labeled as \textit{uncertain}.  

\noindent\textit{Discrepancy Detector}: The \textit{Discrepancy Detector} acts as a second layer for assessing the quality of signals that are labeled as \textit{reliable} in the \textit{Signal Quality} phase by analyzing modality-specific parameters to identify any discrepancies. \textit{Discrepancy Detector} will recognize confounding factors in each modality to identify levels of signal reliability. For instance, for ECG and PPG modalities, we use the range of HR, RR interval (RRi) length, the ratio of Max RRi to min RRi with their thresholds as the parameters to determine discrepancies \cite{signal_quality->ecgppg}. 
As another example, for EDA modality, we use low amplitude and very low and steady tonic level for more than 1 minute as the discrepancy detecting parameter \cite{signal_quality->eda1, signal_quality->eda2}. 


\begin{algorithm}[h]
\small
\setcounter{AlgoLine}{0}
\caption{Signal Monitoring Strategy.} 
\label{alg:sm}
\KwData{$D_M^i, F_M^i, sensorStatus_M^i$}
\KwResult{$Rel_M^i$}
$Rel_M^i \gets RELIABLE$\;
\eIf{$sensorStatus_M^i ==$ DETACHED}{
    $Rel_M^i \gets NOISY$\;
}
{
\eIf{SNR $\leq$ threshold1}{
    $Rel_M^i \gets NOISY$\;
}{
\If{threshold1 $\leq$ SNR $\leq$ threshold2}{
    $Rel_M^i \gets UNCERTAIN$\;
}
}
}


for each modality $D_M^i$\;
\eIf{Rules($D_M^i$) $==$ PASS}{
    $Rel_M^i \gets RELIABLE$\;
}{
    $Rel_M^i \gets UNCERTAIN$\;
}


\end{algorithm}


\noindent\textbf{Adaptive Controller} analyzes inputs from the signal monitoring module to adaptively select features and modalities, and configure sensing and computation models. We store pre-trained models for different combinations of signal modalities and aggregated feature vectors in a pool of models. 
The adaptive controller is designed to holistically and dynamically control the quality of sensing and sense-making, as shown in Algorithm \ref{alg:ac}. 
The adaptive controller considers the reliability of each sensor modality ($Rel_M^i$), the feature vector ($F_M^i$) extracted from \textit{Signal Monitoring}, and the model pool for feature aggregation and model selection. The controller module adaptively selects the features that are relevant for prediction accuracy, and a learning model that is suitable for the updated feature vector using a simple rule based approach. In case a modality is labelled as \textit{uncertain}, we process the modality by dropping some of the less prominent features. If a modality is \textit{noisy}, we drop the input data from the modality entirely. The controller will generate a signal to update the sensing configuration, such that \textit{noisy} sensor modalities can be turned off. Sensor modalities are turned into the idle mode when the noise cannot be mitigated and thereby eliminating the unnecessary computation and communication penalty. 
Finally,  \textit{Model Selection} utilizes the selected features, and the current models available in the \textit{Model Pool} to load a preferable model for the \textit{Inference Engine} taking into consideration the sensing status and situation. Algorithm \ref{alg:ac} is a simple proof-of-concept rule-based policy to show the potentials of this approach. More advanced algorithms (e.g., using reinforcement learning or fuzzy control) can be devised to make intelligent decisions. 

\begin{algorithm}[t]
\setcounter{AlgoLine}{0}
\small
\caption{Adaptive Controller.}
\label{alg:ac}
\KwData{$Rel_M^i, F_M^i, MP$}
\KwResult{Inference Model}
$Sens_M^i \gets ON$\;
\eIf{$Rel_M^i == NOISY$}{
    $Sens_M^i \gets OFF$\;
}{
    \eIf{$Rel_M^i == UNCERTAIN$}{
        $F_M^i \gets$ Choose subset of $F_M^i$
    }{
        $F_M^i \gets F_M^i$
    }
}
\If{$Sens_M^i$ and $F_M^i$}{
    $Inf_M^i \gets$ MP_M^i
}
\end{algorithm}

    
\section{Evaluation}
\vspace{-1mm}
We evaluate the proposed adaptive sensing framework through two case studies of pain assessment \cite{jmir_ecg,jmir_eda, embc_RRpain} and stress monitoring \cite{schmidt2018introducing}. 
Both case study applications require continuous monitoring of multiple modalities including ECG, EMG, PPG, and EDA signals, to detect pain and stress levels. 
Our evaluation platform comprises a sensory node - to collect physiological signals of ECG, EMG, PPG, and EDA from the subject, and an edge node - to execute the inference. 
We implement the proposed method on a ODROID-XU3 with an octa-core Exynos processor as the edge device. 
In the following, we detail the two case studies and evaluate the case study applications' accuracy, energy efficiency, and performance using our AMSER proposed framework through
different scenarios with different types and levels of noisy and unreliable input modalities. 
Further, we compare our AMSER approach against state-of-the-art multi-modal pain monitoring application that use classification of physiological signals \cite{multimodal_classification} as the baseline ideal scenario. 


%

\subsection{Case Studies}
We validate and demonstrate the efficacy of our AMSER approach through two case studies from the affective computing domain: 1) pain assessment and 2) stress monitoring.
\subsubsection{Pain assessment}
State-of-the-art pain assessment tools analyze changes in physiological signals indicating different pain levels. In this work, we evaluate AMSER 
via the pain assessment pre-trained models. These models are trained using UCI\_iHurtDB, a multi-modal dataset from postoperative patients in hospital from 20 patients \cite{jrp}. Data is collected from an eight-channel biopotential acquisition system for EMG and ECG recording and Empatica E4 wristband for PPG and EDA recording. We used the sensor modalities of 500 Hz ECG and EMG, 64 Hz PPG, and 4 Hz EDA during our experiments. We pre-processed each modality and segmented them into 60 second windows. 
For the prediction model to handle potential noisy components in physiological signals, we augmented the raw input training data with true types of noises such as BW and MA. Noisy components induced in raw data are based on the characteristics of each modality and extraction point such as chest, face, or wrist. 
Then, we extract a combination of time domain, frequency domain and automatic features from this window and trained using ML algorithms.
\subsubsection{Stress Monitoring}
The state-of-the-art for stress monitoring presents objective daily stress measurements in everyday settings based on physiological signals \cite{mmstress,  tazarv2021personalized}. In this work, we evaluate our proposed AMSER approach via stress monitoring application based on pre-trained models using WESAD - a multi-modal dataset featuring physiological and motion data recorded from both a wrist- and chest-worn device of 15 subjects during a lab study \cite{schmidt2018introducing}. Sensor modalities used during our experiments are a 700 Hz ECG, 64 Hz PPG, and 4 Hz EDA. We then pre-processed each modality and created 60second window segmentation. We augmented true types of noises such as BW and MA based on the characteristics of each modality. Then, we extracted a combination of time domain, frequency domain and automatic features from this window and trained using ML algorithms.

\subsection{Scenarios}
We evaluated our proposed AMSER approach in 4 different scenarios with different noise component levels for each case study. Table \ref{tab:scenarios} summarizes different scenarios (S1-S4), showing the type of noise induced in each scenario. 
Scenario 1 is the ideal baseline case with no noise components, Scenario 2 has \textit{uncertain} signal components, Scenario 3 has one modality with \textit{noisy} component, and Scenario 4 has two \textit{noisy} modalities.
In Scenario 1, there is no noise component, and thus no modality selection or feature selection is applied. Scenario 2 comprises of baseline wandering noise in addition to the original data for all the modalities. With the presence of noise, our proposed AMSER approach selects specific significant features, reducing the total number of features from the original feature vector. 
In Scenario 3, we add additional motion artifacts on top of one modality while all the modalities still have a baseline wander noise. In this scenario, the noisy modality is completely unreliable. Thus, our proposed method will drop the modality and uses a different model from \textit{Model Pool} with 3 and 2 modalities for pain and stress case studies, respectively. 
Scenario 4 comprises 2 modalities facing severe motion artifact noises. In case of pain assessment, ECG and EMG modalities are noisy with entirely unreliable data. 
Our proposed adaptive sensing will drop both the ECG and EMG modalities and select a learning model with 2 modalities for pain assessment application.
In case of stress monitoring, ECG and EDA modalities are noisy. 
Our proposed adaptive controller drops both the ECG and EDA modalities and selects a learning model with the single PPG modality for stress monitoring.
\begin{table}[]
\vspace{-1mm}
\caption{Experimental Scenarios for Pain and Stress applications. $+BW$, and $+MA$ represent the presence of baseline wandering and motion artifact noise, respectively.}
\label{tab:scenarios}
\resizebox{0.45\textwidth}{!}{
\begin{tabular}{@{}cccccc@{}}
 &  & \multicolumn{1}{c|}{\textbf{ECG}} & \multicolumn{1}{c|}{\textbf{EMG}} & \multicolumn{1}{c|}{\textbf{PPG}} & \textbf{EDA} \\ \midrule
\multicolumn{1}{c|}{\multirow{4}{*}{\rotatebox[origin=c]{90}{\textbf{Pain}}}}   
                        & S1  &  -        &  -      &  -       &  -  \\
\multicolumn{1}{c|}{}   & S2  & +BW       & +BW     & +BW      & +BW \\
\multicolumn{1}{c|}{}   & S3  & +BW       & +BW     & +BW+MA   & +BW \\
\multicolumn{1}{c|}{}   & S4  & +BW+MA    & +BW+MA  & +BW      & +BW \\ \midrule
\multicolumn{1}{c|}{\multirow{4}{*}{\rotatebox[origin=c]{90}{\textbf{Stress}}}} 
                        & S1  &  -        &  N/A    &  -       &  -     \\
\multicolumn{1}{c|}{}   & S2  & +BW       & N/A     & +BW      & +BW    \\
\multicolumn{1}{c|}{}   & S3  & +BW+MA    & N/A     & +BW+MA   & +BW    \\
\multicolumn{1}{c|}{}   & S4  & +BW+MA    & N/A     & +BW      & +BW+MA \\ \bottomrule
\end{tabular}
}
\vspace{-3mm}
\end{table}

We evaluate the accuracy and energy efficiency of both the pain monitoring application (which estimates the pain intensity using 4 physiological signals ECG, EMG, EDA, and PPG), and the stress monitoring application (which calculates the stress level using 3 physiological signals ECG, PPG, and EDA)
using the iHurt\_DB \cite{jrp}. 
We trained different models for each levels of pain intensities, varying in the types of modalities combined for the decision-making system. After pre-processing, we extract a set of unique features from each modality viz., 52 features of ECG, 42 features of EDA, and 42 features of PPG \cite{jmir_ecg, jmir_eda, embc_RRpain, tazarv2021personalized}.

\begin{table}[h]
\caption{Decision made by AMSER during each scenario.}
\label{tab:decision}
\resizebox{0.5\textwidth}{!}{
\begin{tabular}{@{}ccc|c|c|c|c@{}}
 &  & \textbf{ECG} & \textbf{EMG} & \textbf{PPG} & \textbf{EDA} & \textbf{Feature Vol (\%)}\\ \midrule
\multicolumn{1}{c|}{\multirow{4}{*}{\rotatebox[origin=c]{90}{\textbf{Pain}}}} 
& S1 & \checkmark  & \checkmark & \checkmark & \checkmark & 100\\
\multicolumn{1}{c|}{}    & S2 & \checkmark  & \checkmark & \checkmark & \checkmark & 63.8\\
\multicolumn{1}{c|}{}    & S3 & \checkmark  & $\times$   & \checkmark & \checkmark & 35.8\\
\multicolumn{1}{c|}{}    & S4 & $\times$    & $\times$   & \checkmark & \checkmark & 31.3\\
    \midrule
\multicolumn{1}{c|}{\multirow{4}{*}{\rotatebox[origin=c]{90}{\textbf{Stress}}}}  
& S1 & \checkmark  & N/A & \checkmark & \checkmark & 100\\
\multicolumn{1}{c|}{}    & S2 & \checkmark  & N/A & \checkmark & \checkmark & 75.4\\
\multicolumn{1}{c|}{}    & S3 & \checkmark  & N/A   & \checkmark & \checkmark & 31.6\\
\multicolumn{1}{c|}{}    & S4 & $\times$    & N/A   & \checkmark & \checkmark & 14\\ \bottomrule
\end{tabular}
}
\vspace{-3mm}
\end{table}

\subsection{Accuracy Evaluation}
\begin{figure}[t]
\pgfplotsset{every x tick label/.append style={font=\tiny, yshift=0.5ex}}
\pgfplotsset{every y tick label/.append style={font=\tiny, xshift=0.5ex}}
\usetikzlibrary{patterns}
\pgfplotsset{compat=1.11,
	/pgfplots/ybar legend/.style={
		/pgfplots/legend image code/.code={%
			\draw[##1,/tikz/.cd,yshift=-1mm]
			(5mm,2mm) rectangle (2pt,0.2em);},
	},
}
\begin{subfigure}[t]{0.23\textwidth}
\begin{tikzpicture}
\begin{axis}[
ybar=2pt,
grid=major,
enlarge x limits={abs=0.8},
ymin=0,
ymax=100,
height = 4.5cm,
bar width=5pt,
ylabel={\small Accuracy (\%)},
xticklabel style={rotate=0, font=\small},
yticklabel style={rotate=0, font=\small},
xtick = data,
ylabel near ticks,
table/header=false,
every node near coord/.append style={font=\tiny},
table/row sep=\\,
xticklabels from table={
	S1\\
	S2\\
	S3\\
	S4\\
}{[index]0},
legend columns=2,
enlarge y limits={value=.1,upper},
legend style={at={(1,1.05)},anchor=south, font=\normalsize},
]
\legend{AMSER, Baseline}
\addplot [draw=black,draw=black,fill=gray] table[x expr=\coordindex,y index=0]{72.97\\68.48\\66.47\\58.67\\}; 
\addplot [draw=black,draw=black,fill=black] table[x expr=\coordindex,y index=0]{72.97\\46.54\\41.6\\45.4\\}; 

\pgfplotsinvokeforeach{0,1,2,3}{\coordinate(l#1)at(axis cs:#1,0);}
\end{axis}
\end{tikzpicture}
\caption{\textbf{Pain}}
\end{subfigure}
\hspace{5mm}
\begin{subfigure}[t]{0.23\textwidth}
\begin{tikzpicture}
\begin{axis}[
ybar=2pt,
grid=major,
enlarge x limits={abs=0.8},
ymin=0,
ymax=100,
height = 4.5cm,
bar width=5pt,
xticklabel style={rotate=0, font=\small},
yticklabels={,,},
xtick = data,
ylabel near ticks,
table/header=false,
every node near coord/.append style={font=\tiny},
table/row sep=\\,
xticklabels from table={
	S1\\
	S2\\
	S3\\
	S4\\
}{[index]0},
legend columns=4,
enlarge y limits={value=.1,upper},
]
\addplot [draw=black,draw=black,fill=gray] table[x expr=\coordindex,y index=0]{67.07\\66.29\\58.96\\53.99\\}; 
\addplot [draw=black,draw=black,fill=black] table[x expr=\coordindex,y index=0]{67.07\\49.6\\48.4\\43.2\\}; 

\pgfplotsinvokeforeach{0,1,2,3}{\coordinate(l#1)at(axis cs:#1,0);}
\end{axis}
\end{tikzpicture}
\caption{\textbf{Stress}}
\end{subfigure}
\caption{Accuracy analysis for AMSER vs. Baseline \cite{multimodal_classification} for Pain and Stress applications. }
\vspace{-6mm}
\label{fig:acc}
\end{figure}
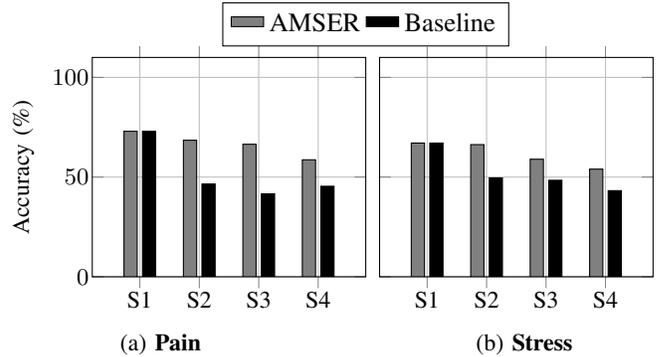
We compare the accuracy achieved by the model in each scenario with noisy components (S2-S4) against the baseline ideal scenario $S1$ with no adaptive modality and feature selection. Figure~\ref{fig:acc} shows the accuracy (in \%) for pain assessment and stress monitoring case studies under different scenarios. 
Our proposed adaptive multimodal sensing technique provides an improved accuracy in each of the scenarios for both the case studies, with a maximum gain of 22\% in S2 for pain assessment. 
In the presence of a lower level of noise i.e., \textit{uncertain} signal, our proposed AMSER approach still utilizes the \textit{uncertain} modality with selected features, instead of either dropping or holding the entire feature sets of the noisy modality. 
This approach leads to a 22\% and 17\% of accuracy improvement for pain and stress applications respectively, in comparison with the baseline model which uses all the features from the noisy modality. 
With a higher level noise i.e., \textit{noisy} modality in scenario $S3$, our proposed AMSER approach leads to 15\% and 10\% accuracy improvement for pain and stress applications, respectively, compared to the baseline which uses all the modalities. 
Lastly, scenario $S4$ shows that dropping 2 noisy modalities results in better accuracy (13\% and 11\% improvement for pain and stress applications, respectively), as compared to the baseline using noisy modalities. 
\subsection{Efficiency and Performance Evaluation}
\begin{figure}[t]
\pgfplotsset{every x tick label/.append style={font=\tiny, yshift=0.5ex}}
\pgfplotsset{every y tick label/.append style={font=\tiny, xshift=0.5ex}}
\usetikzlibrary{patterns}
\pgfplotsset{compat=1.11,
	/pgfplots/ybar legend/.style={
		/pgfplots/legend image code/.code={%
			\draw[##1,/tikz/.cd,yshift=-1mm]
			(5mm,2mm) rectangle (2pt,0.2em);},
	},
}
\begin{subfigure}[t]{0.22\textwidth}
\begin{tikzpicture}
\begin{axis}[
ybar=2pt,
grid=major,
enlarge x limits={abs=0.8},
ymin=0,
ymax=6,
height = 4.5cm,
bar width=5pt,
ylabel={\small \parbox[c]{3cm}{\centering Energy Eff (Baseline=1)}},
xticklabel style={rotate=0, font=\small},
yticklabel style={rotate=0, font=\small},
xtick = data,
ylabel near ticks,
table/header=false,
every node near coord/.append style={font=\tiny},
table/row sep=\\,
xticklabels from table={
	S1\\
	S2\\
	S3\\
	S4\\
}{[index]0},
legend columns=2,
enlarge y limits={value=.1,upper},
legend style={at={(1,1.05)},anchor=south, font=\normalsize},
]
\legend{AMSER-Edge, AMSER-Sensor}
\addplot [draw=black,draw=black,fill=black] table[x expr=\coordindex,y index=0]{1\\1.09\\1.32\\2.19\\}; 
\addplot [draw=black,draw=black,fill=gray] table[x expr=\coordindex,y index=0]{1\\1\\2.63\\5.63\\}; 

\pgfplotsinvokeforeach{0,1,2,3}{\coordinate(l#1)at(axis cs:#1,0);}
\end{axis}
\end{tikzpicture}
\caption{\textbf{Pain}}
\end{subfigure}
\hspace{8mm}
\begin{subfigure}[t]{0.23\textwidth}
\begin{tikzpicture}
\begin{axis}[
ybar=2pt,
grid=major,
enlarge x limits={abs=0.8},
ymin=0,
ymax=6,
height = 4.5cm,
bar width=5pt,
xticklabel style={rotate=0, font=\small},
yticklabels={,,},
xtick = data,
ylabel near ticks,
table/header=false,
table/row sep=\\,
every node near coord/.append style={font=\tiny},
xticklabels from table={
	S1\\
	S2\\
	S3\\
	S4\\
}{[index]0},
legend columns=4,
enlarge y limits={value=.1,upper},
]
\addplot [draw=black,draw=black,fill=black] table[x expr=\coordindex,y index=0]{1\\1.10\\1.31\\3.28\\}; 
\addplot [draw=black,draw=black,fill=gray] table[x expr=\coordindex,y index=0]{1\\1\\1.94\\3.66\\}; 

\pgfplotsinvokeforeach{0,1,2,3}{\coordinate(l#1)at(axis cs:#1,0);}
\end{axis}
\end{tikzpicture}
\caption{\textbf{Stress}}
\end{subfigure}
\vspace{-2mm}
\caption{Energy efficiency analysis of the edge and sensor device for AMSER vs. Baseline \cite{multimodal_classification} for Pain and Stress applications. }
\vspace{-4mm}
\label{fig:energy}
\end{figure}
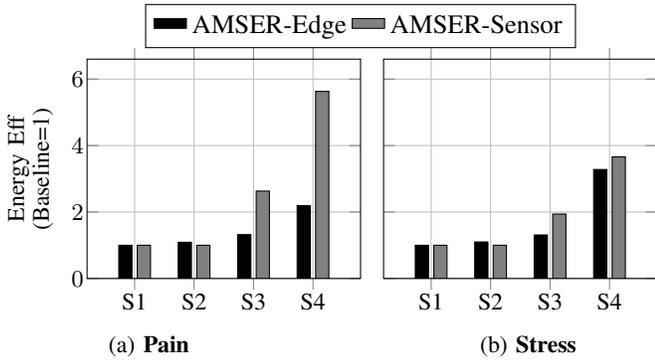

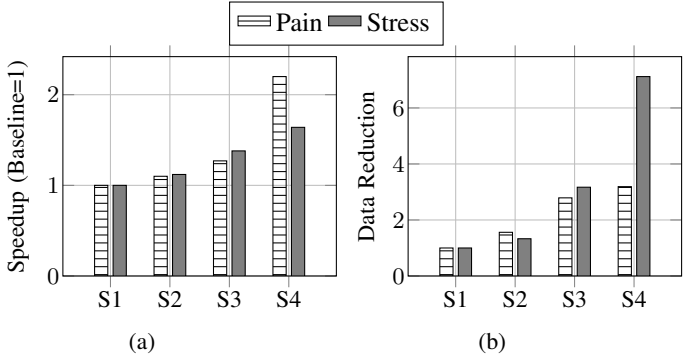
\begin{figure}[t]
\pgfplotsset{every x tick label/.append style={font=\tiny, yshift=0.5ex}}
\pgfplotsset{every y tick label/.append style={font=\tiny, xshift=0.5ex}}
\usetikzlibrary{patterns}
\pgfplotsset{compat=1.11,
	/pgfplots/ybar legend/.style={
		/pgfplots/legend image code/.code={%
			\draw[##1,/tikz/.cd,yshift=-1mm]
			(5mm,2mm) rectangle (2pt,0.2em);},
	},
}
\begin{subfigure}[t]{0.21\textwidth}
\begin{tikzpicture}
\begin{axis}[
ybar=2pt,
grid=major,
enlarge x limits={abs=0.8},
ymin=0,
height = 4.5cm,
bar width=5pt,
ylabel={\small Speedup (Baseline=1)},
xticklabel style={rotate=0, font=\small},
yticklabel style={rotate=0, font=\small},
xtick = data,
ylabel near ticks,
table/header=false,
every node near coord/.append style={font=\tiny},
table/row sep=\\,
xticklabels from table={
	S1\\
	S2\\
	S3\\
	S4\\
}{[index]0},
legend columns=2,
enlarge y limits={value=.1,upper},
legend style={at={(1,1.05)},anchor=south, font=\normalsize},
]
\legend{Pain, Stress}
\addplot [draw=black,pattern=horizontal lines] table[x expr=\coordindex,y index=0]{1\\1.10\\1.27\\2.2\\}; 
\addplot [draw=black,fill=gray] table[x expr=\coordindex,y index=0]{1\\1.12\\1.38\\1.64\\}; 

\pgfplotsinvokeforeach{0,1,2,3}{\coordinate(l#1)at(axis cs:#1,0);}
\end{axis}
\end{tikzpicture}
\caption{}
\end{subfigure}
\hspace{6mm}
\begin{subfigure}[t]{0.21\textwidth}
\begin{tikzpicture}
\begin{axis}[
ybar=2pt,
grid=major,
enlarge x limits={abs=0.8},
ymin=0,
height = 4.5cm,
bar width=5pt,
xticklabel style={rotate=0, font=\small},
yticklabel style={rotate=0, font=\small},
ylabel={\small Data Reduction },
xtick = data,
ylabel near ticks,
table/header=false,
every node near coord/.append style={font=\tiny},
table/row sep=\\,
xticklabels from table={
	S1\\
	S2\\
	S3\\
	S4\\
}{[index]0},
legend columns=4,
enlarge y limits={value=.1,upper},
]
\addplot [draw=black,draw=black,pattern=horizontal lines] table[x expr=\coordindex,y index=0]{1\\1.56\\2.79\\3.19\\}; 
\addplot [draw=black,draw=black,fill=gray] table[x expr=\coordindex,y index=0]{1\\1.33\\3.17\\7.12\\}; 

\pgfplotsinvokeforeach{0,1,2,3}{\coordinate(l#1)at(axis cs:#1,0);}
\end{axis}
\end{tikzpicture}
\caption{}
\end{subfigure}
\vspace{-2mm}
\caption{(a) Performance analysis of the edge device for AMSER vs. Baseline \cite{multimodal_classification} for Pain and Stress applications. (b) Data volume transferred between the sensors and edge for AMSER vs. Baseline.}
\vspace{-6mm}
\label{fig:perfdata}
\end{figure}

We evaluate the efficiency and performance of AMSER in comparison with the baseline. 
Figure~\ref{fig:energy} shows the energy efficiency improvement during four scenarios for edge and sensor devices. We report the results for both Pain and Stress applications. 
Figure~\ref{fig:energy} (a) shows the energy efficiency of the pain assessment application. 
In Scenario $S4$, the energy gain is $2.19\times$ and $5.63\times$ higher than the baseline for the edge and sensors, respectively.
In this case, AMSER provides $13.27\%$ better accuracy by dropping two noisy modalities (See Figure~\ref{fig:acc} and Table~\ref{tab:decision}). 
Figure~\ref{fig:energy} (b) shows the energy efficiency for the stress monitoring application. 
In Scenario $S3$, our evaluation shows that AMSER improves the energy efficiency of the stress monitoring application by $1.32\times$ and $2.63\times$ for the edge and sensor devices, respectively (See Figure~\ref{fig:energy} (b)). 
In this case, AMSER improves the accuracy and performance by $10.56\%$ and $27\%$, respectively (See Figure~\ref{fig:acc}). 
Through the Scenario $S3$ for Stress application, AMSER keeps $31.6\%$ of the features and drops the unreliable features (See Table~\ref{tab:decision}). 
Figure \ref{fig:perfdata} (a) shows the performance gains at the edge device for the pain assessment and stress monitoring applications using our proposed AMSER approach, as compared to the baseline Scenario $S1$. 
In Scenarios $S2$-$S4$, our proposed AMSER approach reduces the number of features and/or drops input data from \textit{noisy} modalities. 
This lowers the total computational effort, leading to an increase in performance, as compared to the baseline. Figure \ref{fig:perfdata} (b) shows the data volume transferred between the sensory devices and edge node using our proposed AMSER approach, in comparison with the baseline. 
Our AMSER approach adaptively selects features and/or drops \textit{noisy} modalities, which reduces the total input data volume significantly. 
For instance, in Scenario $S4$, two noisy modalities are entirely dropped, leading to more than 6$\times$ 
reduction in transmitted data volume. 
Such reduction in data volume to be transmitted further reduces the communication latency and energy expense incurred in transmitting noisy data. 

\section{Conclusion}
We proposed AMSER, a novel adaptive, closed-loop multi-modal sensing framework for energy efficient and resilient eHealth applications. 
AMSER guides sensing and sense-making by monitoring input signal quality and discrepancy detection
via an intelligent control framework that reduces garbage  data to achieve both energy efficiency as well as improved quality. 
We demonstrated the AMSER framework's efficiency with two eHealth case studies on pain assessment and stress detection, achieving up to $2.2\times$ and $5.6\times$ performance and energy gains respectively, compared to the baseline with no optimization, while achieving up to $22\%$ better accuracy. Our future work will focus on devising more intelligent resource management policies for adaptive control and co-optimization of both sensing and sense-making. 


\bibliographystyle{unsrt}
\bibliography{ref}

\end{document}